\documentclass{article}

\usepackage{arxiv}
\usepackage{times}
\usepackage{soul}
\usepackage{url}
\usepackage[hidelinks]{hyperref}
\usepackage[utf8]{inputenc}
\usepackage[small]{caption}
\usepackage{graphicx}
\usepackage{amsmath}
\usepackage{amsthm}
\usepackage{amsfonts}
\usepackage{booktabs}
\usepackage{pgfplots}
\usepackage{subcaption}
\usepackage{algorithm}
\usepackage{algorithmic}
\usepackage{tikz}
\usetikzlibrary{calc,intersections,through,backgrounds}
\usepackage{comment}
\usepackage{array}
\usepackage{booktabs}
\usepackage{natbib}
\usepackage{multirow}
\usepackage{xcolor}
\usepackage{wasysym}
\usepackage{thmtools, thm-restate}
\usepackage{physics}
\usepackage{bbm}
\usepackage{bm}
\usepackage{mathtools}
\usepackage{mathptmx}
\usepackage[shortlabels]{enumitem}
\usepackage{todonotes}
\usepackage[scaled=.90]{helvet}
\usepackage{courier}

\theoremstyle{definition}
\newtheorem{definition}{Definition}
\newtheorem{example}{Example}

\DeclareMathOperator{\area}{Area}
\DeclareMathOperator{\Before}{\textsf{before}}
\DeclareMathOperator{\After}{\textsf{after}}

\DeclareMathOperator{\Start}{\textsf{start}}
\DeclareMathOperator{\End}{\textsf{end}}
\DeclareMathOperator{\Happ}{\textsf{H}}
\DeclareMathOperator{\softplus}{\mathrm{s}_{+}}
\def\N{\mathbb{N}}
\def\R{\mathbb{R}}

\def\L{\mathcal{L}}
\def\N{\mathbb{N}}
\def\R{\mathbb{R}}
\def\E{\mathbb{E}}
\def\I{\mathcal{I}}
\def\T{\mathbb{T}}
\def\B{\mathbb{B}}

\def\bt{\bm t}
\def\G{\mathcal{G}}
\def\I{\mathcal{I}}
\def\G{\mathcal{G}}
\def\I{\mathcal{I}}
\def\before{\ \textsf{bf}\ }
\def\after{\ \textsf{af}\ }
\def\meets{\ \textsf{mt}\ }
\def\overlaps{\ \textsf{ol}\ }
\def\starts{\ \textsf{st}\ }
\def\during{\ \textsf{dr}\ }
\def\finishes{\ \textsf{fin}\ }
\def\contained{\ \textsf{in}\ }
\def\feq{\ \textsf{eq}\ }
\def\Running{\mathrm{Active}}

\def\fuzzyint#1#2#3#4#5#6#7{
  \draw[thick,#1] (#2,0) to (#3,2);
  \draw[thick,#1] (#3,2) to node[below] {#7} (#4,2);
  \draw[thick,#1] (#4,2) to (#5,0);
  \draw[fill=#6,opacity=.2] (#2,0) -- (#3,2) -- (#4,2) -- (#5,0);
  \draw[dashed,#1] (#3,2) -- (#3,0);
  \draw[dashed,#1] (#4,2) -- (#4,0);
  \node[#1] at (#2,-.3) {$#2$}; 
  \node[#1] at (#3,-.3) {$#3$};
  \node[#1] at (#4,-.3) {$#4$};
  \node[#1] at (#5,-.3) {$#5$};     
}

\def\trapz#1#2#3#4#5#6#7#8#9{
  \draw[thick,#1] (#2,#6) to (#3,#7);
  \draw[thick,#1] (#3,#7) to node[below] {#8} (#4,#7);
  \draw[thick,#1] (#4,#7) to (#5,#6);
  \ifthenelse{\boolean{#9}}{
    \draw[dashed,#1] (#3,#7) -- (#3,#6);
    \draw[dashed,#1] (#4,#7) -- (#4,#6);  
  }{}
}

\def\trapzfilled#1#2#3#4#5#6#7#8#9{
  \draw[thick,#1] (#2,#6) to (#3,#7);
  \draw[thick,#1] (#3,#7) to node[below] {#8} (#4,#7);
  \draw[thick,#1] (#4,#7) to (#5,#6);
  \draw[fill=#1,opacity=.2] (#2,#6) -- (#5,#6) -- (#4,#7) -- (#3,#7);
  \ifthenelse{\boolean{#9}}{
    \draw[dashed,#1] (#3,#7) -- (#3,#6);
    \draw[dashed,#1] (#4,#7) -- (#4,#6);  
  }{}
  
}

\def\labeltrapz#1#2#3#4#5#6#7#8{
    \node[#1] at (#2,#6-.3) {$a#8$}; 
    \node[#1] at (#3,#6-.3) {$b#8$};
    \node[#1] at (#4,#6-.3) {$c#8$};
    \node[#1] at (#5,#6-.3) {$d#8$};     
}

\def\trapzlinf#1#2#3#4#5#6#7#8{
    \draw[thick,#1] (#4,#6) to node[below] {#7} (#2,#6);
    \draw[thick,#1] (#2,#6) to (#3,#5);
    \ifthenelse{\boolean{#8}}{
        \draw[dashed,#1] (#2,#6) -- (#2,#5);
    }{} 
  }

\def\trapzrinf#1#2#3#4#5#6#7#8{
    \draw[thick,#1] (#2,#5) to (#3,#6);
    \draw[thick,#1] (#3,#6) to node[below]{#7} (#4,#6);
    \ifthenelse{\boolean{#8}}{
        \draw[dashed,#1] (#3,#6) -- (#3,#5);
    }{} 
  }

\title{Interval Logic Tensor Networks}
\author{%
Samy Badreddine$^1$\quad Gianluca Apriceno$^{2,3}$\quad Andrea Passerini$^{3}$\quad Luciano Serafini$^2$ \\
$^1$Sony AI, Tokyo, Japan \\
$^2$Fondazione Bruno Kessler, Trento, Italy \\
$^3$University of Trento, Trento, Italy \\
\texttt{samy.badreddine@sony.com},
\texttt{apriceno@fbk.eu},\\
\texttt{andrea.passerini@unitn.it},
\texttt{serafini@fbk.eu}
}

\begin{document}

\maketitle

\begin{abstract}
In this paper, we introduce Interval Real Logic (IRL), a two-sorted logic that interprets knowledge such as sequential properties (traces) and event properties using sequences of real-featured data.
We interpret connectives using fuzzy logic, event durations using trapezoidal fuzzy intervals, and fuzzy temporal relations using relationships between the intervals' areas. 
We propose Interval Logic Tensor Networks (ILTN), a neuro-symbolic system that learns by propagating gradients through IRL.
In order to support effective learning, ILTN defines smoothened versions of the fuzzy intervals and temporal relations of IRL using softplus activations.  
We show that ILTN can successfully leverage knowledge expressed in IRL in synthetic tasks that require reasoning about events to predict their fuzzy durations.
Our results show that the system is capable of making events compliant with background temporal knowledge.
\end{abstract}

\section{Introduction}
\label{sec:introduction}

Event detection (ED) from sequences of data is a critical challenge in various fields, including surveillance~\citep{ED:surveillance}, multimedia processing~\citep{xiang2019survey,lai2022event}, and social network analysis~\citep{ED:social_net}. 
Neural network-based architectures have been developed for ED, leveraging various data types such as text, images, social media data, and audio. 
Integrating commonsense and structural knowledge about events and their relationships can significantly enhance machine learning methods for ED. 
For example, in analyzing a soccer match video, the knowledge that a red card shown to a player is typically followed by the player leaving the field can aid in event detection. 
Additionally, knowledge about how simple events compose complex events is also useful for complex event detection. 

Background knowledge has been shown to improve the detection of complex events especially when training data is limited~\citep{yin2020knowledge}. 
Some approaches show how knowledge expressed in first-order logic \citep{vilamala2023deepprobcep,DEEPPROBLOG:video,NS:video} can be exploited for complex event detection. 
Other approaches use temporal logic, such as LTLf, to embed temporal properties in deep-learning architectures processing image sequences~\citep{LTN:Umili_temporal}.

To the best of our knowledge, all existing methods that incorporate background temporal knowledge in event detection adopt a point-wise approach, defining events based on properties that hold (or do not hold) at specific time points during the event's duration. 
However, the knowledge representation and formal ontology literature advocates for event-centric representations, where events are treated as "first-class citizens" with properties that cannot be expressed solely in terms of time-point properties~\citep{kowalski1986logic, TK:Allen_intervals,mueller2008event}. 

The traditional perspective of event representation characterizes events as crisp entities and represents the duration of an event, which is the time span during which it occurs, as a convex subset of integers or real numbers. 
However, this approach does not account for events that have smooth beginnings or endings, such as a snowfall. 
Furthermore, even crisp events can benefit from fuzzy semantics in representing relations between them. 
For example, the statement "Darwin (1809-1882) lived before Einstein (1879-1955)" is not as true as "March 13 comes before March 14," but it is also not entirely false. 
To address this limitation, knowledge representation formalisms have been proposed for fuzzy intervals and fuzzy relations between them \citep{ohlbach2004relations,schockaert2008fuzzifying}.

This paper introduces a novel logical framework that enables the specification of dynamically changing propositions, as well as properties and relations between events. 
We refer to this framework as Interval Real Logic, which is an extension of Real Logic \citep{LTN}. 
This logic is designed to capture knowledge properties and relations between objects that evolve over time, including properties and relations between events. 
Interval Real Logic is interpreted in the domain of real-data sequences, where objects are associated with trajectories, and events are associated with the objects that participate in the event, as well as the temporal interval during which the event occurs.

In addition, the paper introduces the differentiable implementation of Interval Real Logic in a neuro-symbolic architecture, Interval Logic Tensor Networks (ILTN), to detect events from data sequences using background knowledge expressed in Interval Real Logic. 
To effectively propagate gradients through the logic, we propose modified trapezoidal fuzzy membership functions and temporal relations for fuzzy intervals that overcome vanishing gradient issues. 
We present a prototype implementation of ILTN and conduct basic experiments that yield promising and positive results.

The rest of the paper is organised as follows: 
Section \ref{sec:related_work} presents related work on fuzzy temporal knowledge and neuro-symbolic approaches for event detection.
Section \ref{sec:interval_real_logic} defines the language and the semantics of ILTN.
Section \ref{sec:fuzzy_intervals_and_relations} defines fuzzy trapezoidal intervals and their temporal relations.
In Section \ref{sec:architecture}, the neural architecture used to predict fuzzy events is described.
In section \ref{sec:experiments}, the results on artificial experiments are discussed. 
Finally, in Section \ref{sec:conclusion} conclusions are drawn and directions for future works are briefly outlined.


\section{Related Work}
\label{sec:related_work}

Modeling and reasoning about temporal knowledge is a well-studied problem \citep{TK:Kahn, TK:Allen_intervals, TK:Allen_theory_time, TK:Jong}. Temporal logics like Linear Temporal Logic (LTL) \citep{TK:LTL} and Computational Tree Logic (CTL) \citep{TK:CTL} assume that the underlying (temporal) information is crisp, and do not consider that the knowledge may be characterized by vagueness and uncertainty. 
Following the seminal work of \citet{FZ:Zadeh} on fuzzy sets, different works have been proposed to model both vagueness and uncertainty of temporal knowledge when this is expressed in terms of events and their relations via a fuzzy interval-based temporal model ~\citep{FZ:Dubois, FZ:Nagypal, ohlbach2004relations, FZ:Schockaert}.
These works however are not capable of processing low level information in an efficient way, and do not consider any learning. Indeed, fuzzy event recognition applications~\citep{FZ_APP:Kapitanova, FZ_APP:Dima, FZ_APP:Muduli}  simply rely on a (fuzzy) rule-based decision system.

Recently, neuro-symbolic approaches \citep{NS:Pascal_book}, which integrate sub-symbolic and symbolic reasoning and allow to effectively integrate learning and reasoning, have been applied in the context of event recognition. A common solutions consists in introducing a symbolic layer refining the output of a pre-trained neural network~\citep{YOLO_EC_1, YOLO_EC_2, DEEPCEP, PROXY_MODELS_VIOLENCE, IoT_SECURITY}. 
In \citep{NEUROPLEX}, the symbolic layer is replaced by a neural network trained via knowledge distillation to emulate symbolic reasoning. The drawback is that this "neuro-symbolic" layer has to be re-trained from scratch even for a slight change of the knowledge. More recently, fully end-to-end differentiable neuro-symbolic architectures have been proposed, by encoding temporal reasoning primitives into existing frameworks like DeepProbLog~\citep{DEEPPROBLOG:sound,DEEPPROBLOG:video} or Learning Modulo Theories~\citep{NS:video}. However, all these approaches reason in terms of time points, making them incapable of fully expressing the properties of temporal events. The solution we propose here aims to overcome these limitations by directly focusing on temporal intervals. 

LTN \citep{LTN} is an end-to-end neuro-symbolic approach based on fuzzy logic where prior domain knowledge is expressed in terms of Real Logic formulas and interpreted using fuzzy logic semantics. LTN has been applied successfully to solve structured tasks like semantic image interpretation \citep{LTN_SII} and to improve state of the art object classifiers \citep{LTN:FasterLTN}. 
A first temporal extension of LTN has been proposed by \citet{LTN:Umili_temporal}, where Linear Temporal Logic over
finite traces (LTLf) formulas are translated to fuzzy deterministic automaton and applied to solve a sequence classification task. However, as for the other previously mentioned neuro-symbolic approaches, LTLf reasons in terms of time points and thus shares their limitations. By extending LTN to deal with (fuzzy) interval logic primitives we aim to allow them to effectively and efficiently process temporal sequences towards complex event recognition.

\section{Interval Real Logic}
\label{sec:interval_real_logic}

Let $\L_t$ be a first-order language that includes terms referring to the \emph{trajectories} of objects over time. 
The syntax for terms and formulas in $\L_t$ follows the standard syntax of first-order logic.

Similarly, let $\L_e$ be a first-order language, referred to as the \emph{language of events}, which includes a set of symbols $e_1, e_2, \dots$ each associated with an arity $m\geq 0$. 
The terms of $\L_e$ are expressed in the form $e(t_1,\dots,t_n)$ if $e$ has arity $m$ and $t_i$'s are terms in $\L_t$. 
Intuitively, $e(t_1,\dots,t_m)$ denotes an event that involves $t_1,\dots,t_m$ as participants. 
Additionally, we assume that $\L_e$ contains the set of binary predicates that correspond to binary relations between events.

\begin{example}
Suppose that we want to describe the events that happen when 
two particles move in a 2D space as shown
in Figure~\ref{fig:trajectories-a-b}.

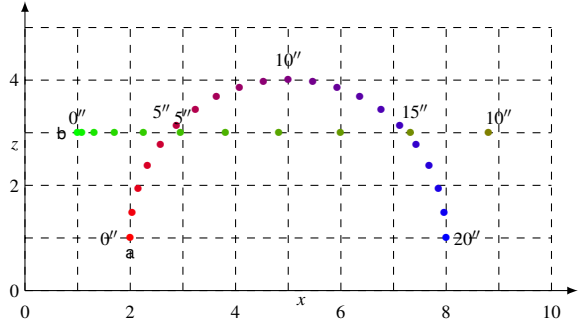
\begin{figure}[h]
\begin{center}
  \begin{tikzpicture}[scale=.7,every node/.style={scale=.7}]
    \def\n{20}
    \draw[dashed] (-5,-1) grid +(10,5);
    \draw[-latex] (-5,-1) to node[below]{$x$} +(10.5,0);
    \draw[-latex] (-5,-1) to node[left]{$z$} +(0,5.5);
    \foreach \i in {0,2,...,10}
    \node at (\i-5,-1.4) {$\i$};
    \foreach \i in {0,2,...,4}
    \node at (-5.2,\i-1) {$\i$};
    \foreach \a in {0,...,\n}{ \pgfmathparse{5*\a};
      \node[red!\pgfmathresult!blue] at (\a*180/\n:3) {$\bullet$};};
    \foreach \a in {0,...,10}{ \pgfmathparse{5*\a};
      \node[red!\pgfmathresult!green] at (-4+.078*\a*\a,2)
      {$\bullet$};};
  \foreach \a in {0,...,4}{
      \node at (180-\a*5*180/\n:3.4cm) {$\pgfmathparse{int(5*\a)}\pgfmathresult''$};
    }
    \foreach \a in {0,...,2}{
        \node at (-4+2*\a*\a,2.3) {$\pgfmathparse{int(5*\a)}\pgfmathresult''$};}
  
    \node at (-3,-.3) {\textsf{a}};
    \node at (-4.3,2) {\textsf{b}};    
  \end{tikzpicture}
\end{center}

\caption{\label{fig:trajectories-a-b} Trajectories of two particles
  in a 2D space. Several events happen over time. For example, around time $5''$,
  the two particles intersect. From time $0''$ to $10''$,
  particle $a$ rises whereas particle $b$ accelerates from left to right.
 Additionally, over the whole trajectory, particle $a$ is doing a jump while changing color. }
\end{figure}
$\L_t$ and $\L_e$ are used conjointly to describe Figure~\ref{fig:trajectories-a-b}.
In $\L_t$, the two particles are denoted by constants \textsf{a} and \textsf{b}. 
Unary predicates such as $\textsf{blue}$, $\textsf{red}$, and $\textsf{violet}$ are included to describe the particles' colors over time. 
The atomic formula $\textsf{blue(a)}$ expresses that $\textsf{a}$ is blue, with its truth value being time-dependent. 
To describe the proximity of the particles, $\L_t$ uses the binary predicate $\textsf{close}$, and $\textsf{close(a,b)}$ is true around time $5''$ and false otherwise.

In $\L_e$, event symbols are used to describe the events in the figure. 
For example, $\textsf{e0(a)}$ can denote the jump of particle $\textsf{a}$, $\textsf{e1(a)}$ can denote the color change of $\textsf{a}$, and $\textsf{e2(a,b)}$ can denote the event of $\textsf{a}$ and $\textsf{b}$ intersecting. 
Predicates and functions on events are also included in $\L_e$. 
For example, unary predicates on events can be used to specify their types, as in the formula $\textsf{Jump(e0(a))}$, which states that $\textsf{e0(a)}$ is of type jump, and $\textsf{ChangeOfColor(e1(a))}$, which states that $\textsf{e1(a)}$ is of type color change.
\end{example}

We require that $\L_e$ contains the unary functions on
events and the binary relations of events shown in
Table~\ref{tab:function-and-relations-on-events}. 
In the table and in the rest of the paper we use $\epsilon$ (possibly with indices) to denote an event term $e(t_1,\dots,t_m)$.

\begin{table}[h]
\centering
\begin{tabular}{ll} \hline
\multicolumn{2}{c}{\bf Function symbols of $\L_e$} \\ \hline 
 $\Before(\epsilon)$ &  what happens before the starting of $\epsilon$\\ 
 $\After(\epsilon)$ & what happens after the end of $\epsilon$ \\ 
 $\Start(\epsilon)$ & the starting of $\epsilon$ \\
 $\End(\epsilon)$ & the end of $\epsilon$ \\  
 $[i,j]$  & for $i\leq j\in\N$ \\
\hline 
\multicolumn{2}{c}{\bf Allen's predicate symbols of $\L_e$} \\ \hline 
$\epsilon_1\before \epsilon_2$ & $\epsilon_1$ happens before $\epsilon_2$ \\
$\epsilon_1\after \epsilon_2$ & $\epsilon_1$ happens after $\epsilon_2$ \\
$\epsilon_1\meets \epsilon_2$ & $\epsilon_2$ happens immediately after $\epsilon_1$ \\
$\epsilon_1\overlaps \epsilon_2$ & the end of $\epsilon_1$ overlaps the start of $\epsilon_2$\\
$\epsilon_1\starts \epsilon_2$ & $\epsilon_1$ is a starting part of  $\epsilon_2$ \\
$\epsilon_1\during \epsilon_2$ & $\epsilon_1$ happens during  $\epsilon_2$ \\
$\epsilon_1\finishes \epsilon_2$ & $\epsilon_1$ is an ending part of  $\epsilon_2$ \\
$\epsilon_1\feq\epsilon_2$ & $\epsilon_1$ is equal to  $\epsilon_2$ \\
  \hline 
\multicolumn{2}{c}{\bf Other predicate symbols of $\L_e$} \\ \hline 
$\Happ(\epsilon)$ & the event $\epsilon$ actually happened  \\ 
$\epsilon_1 \contained \epsilon_2$ & $\epsilon_1$ is contained in
                                     $\epsilon_2$ \\
\end{tabular}
\caption{Basic functions and relations on events}
\label{tab:function-and-relations-on-events}
\end{table}

Finally, $\L_t$ contains a unary predicate $\Running$ that takes as
input an event term. Intuitively, $\Running(\epsilon)$ returns
for every time step of the sequence if the event is running or not. 
\begin{example}
  Following are some examples of formulas in $\L_t$ and $\L_e$.
  The atomic formula
  \begin{align*}
    \textsf{Sunny(weather)}\rightarrow \textsf{Happy(John)}
  \end{align*}
  is an example of a $\L_t$ formula that states that John is happy whenever it is sunny. 
  This formula is evaluated along two traces, one for the weather and one for John, and can take different values at different time points. 
  
  The following $\L_e$ formula
  \begin{align*}
    \Happ(\textsf{e1(John,Mary)}) \wedge \textsf{Meeting(e1(John,Mary))}
  \end{align*}
  states that a meeting between John and Mary happened. 
  
  The $\L_t$ formula
    \begin{align*}
      \Running(\textsf{e1(John,Mary)})\rightarrow\textsf{Happy(John)}
      \wedge \textsf{Happy(Mary)}
  \end{align*}
  expresses that during the meeting between John and Mary, they were both happy.

  The $\L_e$ formula 
  \begin{align*}
    \forall_t x,y.\textsf{Meeting}(\textsf{e1}(x,y))\rightarrow \textsf{e1}(x,y) = \textsf{e1}(y,x)
  \end{align*}
  states that in a meeting event, the roles of the participants are symmetric. 
  Notice that the quantification is on trace variables (not on the events). 
  This is highlighted by the index $t$ of the universal quantifier.
  
  Finally, the $\L_e$ formula 
  \begin{align*}
    \forall_e x.\textsf{Meeting}(x) \rightarrow \exists_e y.
    \textsf{PrepareAgenda}(y) \wedge x \before y 
  \end{align*}
  expresses that before every meeting there should be an event that is the
  preparation of the agenda. In this case, the quantification is on
  event variables, indicated by the index $e$ of the quantifier. 
\end{example}

The semantics of the trace logic $\L_t$ and the event-based logic $\L_e$ are defined in the context of a linear discrete structure, which models the progression of time. 
We use the natural numbers $\N$ with the standard order $<$ as the reference structure for time.

\subsection{Trace Semantics}
\label{s:trace_semantics}
In $\L_t$, terms are interpreted as (possibly infinite) sequences of data, called \emph{trajectories}. 
For each time point $i\in\N$, an $\L_t$ term corresponds to a feature vector in $\R^n$. 
Specifically, a trajectory is a function $\bt:\N\rightarrow\R^n$ that assigns a feature vector in $\R^n$ to every time point.
We denote the set of trajectories with features in $\R^n$ as $\T^n$. 
Trace variables in $\L_t$ refer to variables of individuals and are associated with batches of traces. 
Constants and closed terms (i.e., terms without variables) in $\L_t$ are interpreted as single traces.

Formulas in $\L_t$ are evaluated at all time instants. 
For every time $i\in\N$, an $\L_t$ formula is associated with a truth value in the range $[0,1]$ that represents the level of truth of the formula at that time. 
As a result, an $\L_t$ formula is interpreted as a sequence of truth values, which we refer to as a function from $\N$ to $[0,1]$. 
The set of such functions is denoted as $\B$.

The formal definition of the semantics for $\L_t$ is based on a \emph{grounding} function $\G$ that must satisfy the following conditions:
\begin{itemize}
\item for every variable $x$ in $\L_t$, $\G(x)\in(\T^n)^b$ is a batch of trajectory with the integer size $b\geq1$,
\item for every constant $c\in\L_{t}$, $\G(c)\in\T^n$ is a single trajectory,
\item for every function $f\in\L_t$, with arity equal to $m$,
    $\G(f): \T^{n_1}\times\dots\times\T^{n_m}\rightarrow\T^n$, that is $\G(f)$ maps to a function that takes $m$ input trajectories and returns a trajectory,
\item for every predicate $p\in\L_t$, with arity equal to $m$,
  $\G(p): \T^{n_1}\times\dots\times\T^{n_m}\rightarrow\B$, that is $\G(p)$ maps to a function that takes $m$ input trajectories and outputs a function from time points to truth values in $[0,1]$.
\end{itemize}

Propositional connectives are interpreted according to fuzzy logic semantics which is applied point-wise. 
For example, if $\phi$
and $\psi$ are $\L_t$-formulas, then
$\G(\phi\wedge\psi)=T(\G(\phi),\G(\psi))=\{T(\G_i(\phi),\G_i(\phi))\}_{i\in\N}$,
where $T$ is a t-norm such as the product t-norm. Universal and existential quantifiers are
interpreted as aggregation operators. For example, $\G(\forall x\phi(x))=
\{\prod_{1\leq j\leq b}\G_i(\phi(\G_j(x)))\}_{i\in\N}$. 

Finally, we allow a special predicate that maps from events to $\L_t$:
\begin{itemize}
    \item for every event $\epsilon$,
      $\G(\Running(\epsilon)) : \E^{\bm n} \rightarrow \B$; 
      $i \mapsto T(\I(\epsilon)(i), \Happ(\epsilon))$ where
      $T$ is a t-norm. The functions $\I$ and $\Happ$, as well as the notation $\E^{\bm n}$, are defined in Section \ref{s:event_semantics}. 
      Intuitively, $\Running(\epsilon)$ maps an event to a boolean trajectory that states \emph{when} and \emph{if} the event happens at each timepoint of the trajectory.
\end{itemize}

\subsection{Event Semantics}
\label{s:event_semantics}
An event is seen as a potentially infinite sequence of data, (i.e., a trajectory) and a mask that indicates the duration of the event. 
Formally, an event $\epsilon\in\T^{n_1}\times\dots\times\T^{n_m}\times\B$ consists of $m$ traces, which are the traces of the objects involved in the event $\epsilon$, and a boolean trace that indicates when the event is active. 
Specifically, let $\I(\epsilon)$ denote the boolean trace $\B$ that is the activation sequence of $\epsilon$. 
If $\bm n = (n_1,\dots,n_m)$, we denote $\E^{\bm n}$ as $\T^{n_1}\times\dots\times\T^{n_m}\times\B$, which represents the space of events involving $m$ objects, each with features in $\R^{n_i}$. 
The formal semantics of $\L_e$ is defined in reference to the definition of an event provided in \cite{guarino2022events} and is given in terms of a function $\G$ that satisfies the following restrictions.
\begin{itemize}
\item For every event term $e(t_1,\dots,t_m)$, $\G(e(t_1,\dots,t_m))\in\E^{\bm n}$ where $\bm n=(n_1,\dots,n_m)$ and $\G(t_i)\in\R^{n_i}$ for $1\leq i \leq m$,
\item for every $[i,j]\in\N$, $\G(i)=\{\mathbbm{1}_{n\in[i,j]}\}_{n\in\N}$,
\item for every function symbol $f\in\L_e$, with arity equal to $m$, 
$\G(f):(\E^{\bm n_1}\times\dots\times\E^{\bm n_m})\rightarrow\E^{\bm n_1\cdots\bm n_m}$,
\item for every predicate symbol $p\in\L_e$, with arity equal to $m$,
    $\G(p):(\E^{\bm n_1}\times\dots\times\E^{\bm n_m})\rightarrow[0,1]$.
\end{itemize}
Connectives in $\L_e$ are interpreted using fuzzy semantics. 
For example, $\G(\phi_1\wedge\phi_2)=T(\G(\phi_1),\G(\phi_2))$ where $T$ is a t-norm. 
Quantifiers of events are interpreted by aggregation functions. 

\begin{example}
  The first segment of the grounding $\G$ of the particle $a$ of
  Figure~\ref{fig:trajectories-a-b} is shown in Figure~\ref{fig:g-trajectories-a-b}
\begin{figure}
\scriptsize
\begin{align*}
  \begin{array}{|c|c|c|c|c|c|c|c|c|c|c|c|c|c|c|c|}
    \hline 
    \rotatebox{90}{$i$} &&&  0 & 1 & 2 & 3 & 4 & 5 & 6  \\ \hline
\multirow{11}{*}{\rotatebox{90}{$\G(\textsf{e2(a,b)})$}} &
\multirow{5}{*}{$\G(\textsf{a})$}
    & x & 2.0 & 2.04 & 2.15 & 2.33 & 2.57 & 2.88 & 3.24 \\
&& y & 1.0 & 1.47 & 1.93 & 2.36 & 2.76 & 3.12 & 3.43 \\
&& r & 1.0 & 0.95 & 0.9 & 0.85 & 0.8 & 0.75 & 0.7 \\
&& g & 0.0 & 0.0 & 0.0 & 0.0 & 0.0 & 0.0 & 0.0 \\
&& b & 0.0 & 0.05 & 0.1 & 0.15 & 0.2 & 0.25 & 0.3
\\ \cline{2-10}
& \multirow{5}{*}{$\G(\textsf{b})$}
& x & 1.0 & 1.02 & 1.08 & 1.18 & 1.32 & 1.5 & 1.72 \\
&& y & 4.0 & 4.0 & 4.0 & 4.0 & 4.0 & 4.0 & 4.0 \\
&& r &  1.0 & 0.95 & 0.9 & 0.85 & 0.8 & 0.75 & 0.7 \\
&& g & 0.0 & 0.05 & 0.1 & 0.15 & 0.2 & 0.25 & 0.3 \\
&& b& 0.0 & 0.0 & 0.0 & 0.0 & 0.0 & 0.0 & 0.0 \\
\cline{2-10}
& \I &  & 
\tikz\draw[fill=red] (0,0) rectangle (.3,0.0); & 
\tikz\draw[fill=red] (0,0) rectangle (.3,0.1); & 
\tikz\draw[fill=red] (0,0) rectangle (.3,0.3); & 
\tikz\draw[fill=red] (0,0) rectangle (.3,0.45); & 
\tikz\draw[fill=red] (0,0) rectangle (.3,0.5); & 
\tikz\draw[fill=red] (0,0) rectangle (.3,0.25); & 
\tikz\draw[fill=red] (0,0) rectangle (.3,0.0); \\ \hline 
\end{array}
\end{align*}
\caption{Grounding traces and events for the particle example. $\textsf{e2(a,b)}$ is the event of the two particles intersecting.}
\label{fig:g-trajectories-a-b}
\end{figure}
\end{example}

Examples of function symbols from $\L_e$ include $\Before(\epsilon)$ or $\Start(\epsilon)$, 
whereas examples of predicate symbols of $\L_e$ include $\Happ(\epsilon)$ (unary symbol) and $\epsilon_1 \before \epsilon_2$ (binary symbol in infix notation). 
These symbols are intuitively described in Table \ref{tab:function-and-relations-on-events}.
Their actual grounding is discussed in Section \ref{sec:fuzzy_intervals_and_relations}.
\section{Fuzzy Intervals and Relations}
\label{sec:fuzzy_intervals_and_relations}
As previously mentioned, $\I(\epsilon)$ denotes the activation sequence of an event in $\B$. 
Notice that $\I(\epsilon)$ is a fuzzy subset of $\N$. 
A requirement imposed in \cite{guarino2022events} is that $\I(\epsilon)$ must be an interval, i.e., a convex subset of time points. 
However, in this paper, we consider the fact that such an interval is a fuzzy interval. 
Therefore, we propose imposing constraints on the shape of such a subset to be a \emph{trapezoidal fuzzy number}~\citep{abbasbandy2009new}.

\begin{definition}[Fuzzy interval]
A fuzzy interval is a fuzzy set $I:\R \rightarrow[0,1]$ 
such that there exists $a\leq b\leq c\leq d\in \R$. 
\begin{align}
  \label{eq:fuzzy-interval}
  I(x) & =
         \begin{cases}
            \frac{x-a}{b-a} & \mbox{if $x \in (a,b)$,} \\
            1 & \mbox{if $x \in [b,c]$,} \\
            \frac{x-d}{c-d} & \mbox{if $x \in (c,d)$,} \\
            0 & \mbox{otherwise.} 
         \end{cases}
\end{align}
We also allow special cases of semi-infinite intervals, which we use to define the $\Before$ and $\After$ operators in Section \ref{s:basic_operators}.
\begin{description}
    \item[Left-infinity] A left-infinite fuzzy interval is characterized by the parameters $I=\{x \mid -\infty,-\infty,c,d\}$.
    \item[Right-infinity] A right-infinite fuzzy interval is characterized by the parameters $I=\{x\mid a,b,+\infty,+\infty \}$. 
\end{description}

\end{definition}

\begin{example}

\begin{center}
\begin{tikzpicture}[y=20]
    \draw[-latex] (0,0) to node[below] {$x$} (8,0);
    \draw[-latex] (0,0) to (0,2.2);
    \node at (-.3,2) {$1$};
    \node at (-.3,0) {$0$};      
    \fuzzyint{red}1257{red}{$I\{x\mid 1,2,5,7\}$};
\end{tikzpicture}
\end{center}
\end{example}

With this restriction, we impose that the activation 
function of every event $\I(\epsilon)$ is such that 
there is a trapezoidal fuzzy interval $I=\{x\mid a,b,c,d\}$ such that 
$\I_n(\epsilon)=I(n)$ for every $n\in\N$.
For ease of notation, in the rest of the paper, 
we will commonly denote an interval simply by its four parameters $I=(a,b,c,d)$.

\subsection{Basic Operations on Fuzzy Intervals}
\label{s:basic_operators}
To provide the semantics for the functions and relations of $\L_e$, we
first define a set of basic operations on fuzzy intervals.  Our
operations are inspired by \cite{ohlbach2004relations} who
defines such operations on any convex and non-convex interval.  We
simply specialize them on trapezoidal fuzzy intervals.  

\subsubsection{Duration}
The duration of a trapezoidal fuzzy interval $A=(a,b,c,d)$, denoted by
$|A|$ is equal to $\int_{-\infty}^{+\infty}A(x) dx$.
If $A$ is finite then $|A| = \frac{(c-b)+(d-a)}{2}$, 
otherwise $|A|=\infty$. 

\subsubsection{Before and After}
If $A=(a,b,c,d)$ then $\Before(A) = (-\infty,-\infty,a,b)$ and 
$\After(A) = (c,d,+\infty,+\infty)$ and

\begin{center}
    \begin{tikzpicture}[y=20,x=20]
      \draw[latex-latex] (-1,0) to (9,0);
      \node at (-1.35,0) {$0$};
      \node at (-1.35,1.5) {$1$};
      \trapz{black}{2}{3}{4}{6}{0}{1.5}{$A$}{true}
      \trapzlinf{red}{2}{3}{0}{0}{1.5}{$\Before(A)$}{true}
      \trapzrinf{blue}{4}{6}{8}{0}{1.5}{$\After(A)$}{true}
      \draw[dashed,thick,red] (0,1.5) -- (-1,1.5);
      \draw[dashed,thick,blue] (8,1.5) -- (9,1.5);
    \end{tikzpicture}
\end{center}

\subsubsection{Start and End} 
If $A=(a,b,c,d)$ is left-finite then $\Start(A)$ is defined as 
$(\chi-\frac{\delta}{2}, \chi,\chi, \chi+\frac{\delta}{2})$, 
such that $\chi=\frac{a+b}{2}$ and $\delta=\max(\frac{b-a}{2},\delta_{\min})$ 
where $\delta_{\min}$ is a small positive value to account for the crisp case $a=b$.

Similarly, $\End(A)=(\chi-\frac{\delta}{2}, \chi, \chi, \chi+\frac{\delta}{2})$ 
such that $A$ is right-finite, $\chi=\frac{c+d}{2}$ and $\delta=\max(\frac{d-c}{2},\delta_{\min})$.

\begin{center}
    \begin{tikzpicture}[y=20,x=20]
      \draw[latex-latex] (-1,0) to (9,0);
      \node at (-1.35,0) {$0$};
      \node at (-1.35,1.5) {$1$};
      \trapz{black}{2}{3}{4}{6}{0}{1.5}{$A$}{true}
      \trapz{red}{2}{2.5}{2.5}{3}{0}{1.5}{}{true}
        \node[red,above] at (2.5,1.5) {$\Start(A)$};
      \trapz{blue}{4}{5}{5}{6}{0}{1.5}{}{true}
        \node[blue,above] at (5,1.5) {$\End(A)$};
    \end{tikzpicture}
\end{center}

\subsection{Relations between Fuzzy Intervals}
\label{s:relations}
\cite{ohlbach2004relations} defines interval-interval
relations by computing the integral of point-interval relations over
the points in a set.  To avoid the complexity associated with the
integrals, and to be more compliant with Allen's definition in the
crisp case, we define new relations based on simplified containment ratios.
\begin{align*}
    A \contained B &\coloneqq \frac{|A \cap B|}{|A|} \\
    A \feq B &\coloneqq A \contained B \land B \contained A\\
    A \before B &\coloneqq A\contained \Before(B) \\ 
    A \after B &\coloneqq B\contained \After(A) \\ 
    A \meets B &\coloneqq \End(A)\feq\Start(B) \\
    A \starts B &\coloneqq \Start(A) \feq \Start{B} \wedge \End(A)
                  \before \End(B) \\
    A \during B & \coloneq \Start(A) \after \Start(B) \land \End(A)
                  \before \End(B) \\
    A \finishes B &\coloneqq \Start(A) \after \Start{B} \wedge \End(A)
                  \feq \End(B) \\
    A \overlaps B &\coloneqq \Start(A)\before\Start(B) \land \Start(B)
                    \before \End(A) \\ & \ \ \ \ \land
                    \End(A)\before\End(B)
\end{align*}
In these definitions, the temporal relations take precedence over the fuzzy conjunction $\land$.
For general fuzzy intervals, $|A \cap B|$ can be hard to compute.
However, with trapezoidal intervals, the calculation of $|A\cap B|$ 
is derivable analitically by solving simple linear constraints
system. We show how this is done in the following subsection.


\subsection{Area Intersection}
\label{appx:area_intersection}
Let us calculate $\area(A \cap B)$ for any two finite intervals $A = (a,b,c,d)$ and $B=(a',b',c',d')$.
Without loss of generality, suppose that $a \leq a'$.
Developing an explicit formula to compute $\area(A \cap B)$ is not immediate as the shape of $A \cap B$ can be a polygon with a varying number of edges (at most 6). 
For example:
\begin{center}
    \begin{tikzpicture}[y=20,x=20]
      \draw[-latex] (0,0) to (8,0);
      \draw[-latex] (0,0) to (0,2);
      \node at (-.35,0) {$0$};
      \node at (-.35,1.5) {$1$};
      \trapzfilled{red}{0.5}{2.5}{4.5}{5.5}{0}{1.5}{$A$}{true}
      \trapzfilled{blue}{1}{2}{4}{7}{0}{1.5}{$B$}{true}
      \labeltrapz{blue}{1}{2}{4}{7}{0}{1.5}{'}
      \labeltrapz{red}{0.5}{2.5}{4.5}{5.5}{0}{1.5}{}
    \end{tikzpicture}
\end{center}

We propose to first determine the vertices of the shape  $A\cap B$, and then compute the area of the shape using the shoelace formula.

\paragraph{Empty intersection} First, we dismiss the case $d\leq a'$, in which the two intervals do not intersect. $\area(A \cap B) = 0$.

\begin{center}
    \begin{tikzpicture}[y=20,x=20]
      \draw[-latex] (0,0) to (8,0);
      \draw[-latex] (0,0) to (0,2);
      \node at (-.35,0) {$0$};
      \node at (-.35,1.5) {$1$};
      \trapzfilled{red}{0.5}{1}{2}{3.5}{0}{1.5}{$A$}{false}
      \trapzfilled{blue}{4}{5}{6}{7}{0}{1.5}{$B$}{false}
    \end{tikzpicture}
\end{center}

In the rest of the section, we assume that an intersection always exists.

\paragraph{Bottom vertices}
We call bottom vertices of the shape $A \cap B$, the ones on the line $y=0$.
There are always two.
As $a \leq a'$, $(a',0)$ is always a vertex of the shape.
The second vertex is $(\min(d,d'),0)$.

\paragraph{Top vertices}
We call top vertices the ones on the line $y=1$.
There can be zero, one, or two top vertices that delimit $A \cap B$, as shown in the below figures:

\begin{center}
    \begin{tikzpicture}[y=20,x=20]
      \draw[-latex] (0,0) to (8,0);
      \draw[-latex] (0,0) to (0,2);
      \node at (-.35,0) {$0$};
      \node at (-.35,1.5) {$1$};
      \trapzfilled{red}{1}{2}{3}{4}{0}{1.5}{$A$}{false}
      \trapzfilled{blue}{3}{3.5}{5}{6}{0}{1.5}{$B$}{false}
    \end{tikzpicture}
    ~
    \begin{tikzpicture}[y=20,x=20]
      \draw[-latex] (0,0) to (8,0);
      \draw[-latex] (0,0) to (0,2);
      \node at (-.35,0) {$0$};
      \node at (-.35,1.5) {$1$};
      \trapzfilled{red}{1.5}{2.5}{3.5}{4.5}{0}{1.5}{$A$}{false}
      \trapzfilled{blue}{3}{3.5}{5}{6}{0}{1.5}{$B$}{false}
    \end{tikzpicture}
    ~
    \begin{tikzpicture}[y=20,x=20]
      \draw[-latex] (0,0) to (8,0);
      \draw[-latex] (0,0) to (0,2);
      \node at (-.35,0) {$0$};
      \node at (-.35,1.5) {$1$};
      \trapzfilled{red}{1.5}{2.5}{3.5}{4.5}{0}{1.5}{$A$}{false}
      \trapzfilled{blue}{2.5}{3}{4.5}{5.5}{0}{1.5}{$B$}{false}
    \end{tikzpicture}
\end{center}

If $c<b'$ or $b>c'$, there are zero top vertices.

If $b'=c$, the only top vertex is $(c,1)$. 
If $b=c'$, the only top vertex is $(b,1)$.

In other cases, there are always two top vertices $(\max(b,b'),1)$ and $(\min(c,c'),1)$.

\paragraph{Side vertices}
To determine the side vertices that delimit $A \cap B$, we compute the intersection of the lines drawn by the edges of each trapezoid over the whole $xy$ plane.
Then, we keep the intersections where $y \in [0,1]$.
For example, in the below figure, there is only one intersection that defines a side vertex of $A\cap B$:
\begin{center}
    \begin{tikzpicture}[y=25,x=20]
      \draw[-latex] (0,0) to (8,0);
      \draw[-latex] (0,0) to (0,1);
      \node at (-.35,0) {$0$};
      \node at (-.35,1) {$1$};
      \trapzfilled{red}{1}{2}{3}{3.5}{0}{1}{$A$}{false}
      \trapzfilled{blue}{1.5}{3.5}{4}{7}{0}{1}{$B$}{false}
      \draw[dashed, red] (0,-1) to node[very near end, above]{$L_A$} (4,3);
      \draw[dashed, red] (2,3) to node[at start, below]{$R_A$}(4,-1);
      \draw[dashed, blue] (0,-0.75) to node[at end, above]{$L_B$} (5,1.75);
      \draw[dashed, blue] (1.5,1.83) to node[at start, above]{$R_B$} (8,-0.33);
      
    \end{tikzpicture}
\end{center}

Let us denote $L_A \equiv y=\frac{x-a}{b-a}$ the line drawn by the left side of $A$, 
and $R_A \equiv y=\frac{x-d}{c-d}$ the line drawn by the right side of $A$.
Similarly, we have $L_B \equiv y=\frac{x-a'}{b'-a'}$ and $R_B \equiv y=\frac{x-d'}{c'-d'}$ defined on $B$.

We are interested in finding the four intersections $L_A \cap L_B$, $L_A \cap R_B$, $R_A \cap L_B$, and $R_A \cap R_B$.
Each is easy to determine by solving the system of two equations associated with the pair of lines.
For example, $L_A \cap L_B$ is the point $(\frac{ab'-ba'}{a-b+b'-a'},\frac{a-a'}{a-b+b'-a'})$.

Once we have determined the intersections, we keep the ones where $y \in [0,1]$ to define the vertices of $A\cap B$.

Let us cover some of the edge cases about these intersections.
Firstly, any of the edge lines can be vertical if the trapezoid is crisp on that edge.
For example, if $a=b$, $L_A$ is defined by the equation $x=a$. 
Regardless, the method is the same: we simply use this vertical equation in the system of two equations.
Secondly, it is possible that there are no side vertices if some lines are parallel.
For example, $L_A \cap L_B$ gives no solution if $a-b=a'-b'$ (or infinite solutions if the lines are the same). 
In such cases, we ignore the pair of parallel lines.
Finally, it is also possible that a side vertex is a top or bottom vertex if the lines intersect on $y=0$ or $y=1$.

\paragraph{Area calculation}
Once we have determined all the vertices $(x_i,y_i)$ of $A \cap B$, arranged in a counter-clockwise sequence of points, 
we can calculate the area using the shoelace formula:
\begin{equation}
 \area(A \cap B) = \frac 1 2 \sum_{i=1}^n (y_i + y_{i+1})(x_i - x_{i+1})
\end{equation}

\paragraph{Semi-infinite intervals}
We sometimes have to compute the area intersection in cases where $A$ is left-infinite or $B$ is right-infinite (for example, when using the operators $\before$ or $\after$).
However, we can turn these semi-infinite intervals to finite intervals such that the area calculation is unchanged.
If $A$ is left-infinite, we can replace the infinite parameters with any $a \leq a'$ and $b \leq b'$.
Similarly, if $B$ is right-infinite, we can replace the infinite parameters with any $c'\geq c$ and $d' \geq d$.
Doing so, we can reuse the method highlighted above.
\section{Architecture}
\label{sec:architecture}
The main objective of introducing Interval Real Logic (IRL) is to use it to
impose temporal constraints in a neural architecture for Event Detection. Given a
temporal data sequence $\bm u=\{u_i\}_{i=0}^{T}$ we define a neural
architecture, called \emph{Interval Logic Tensor Networks (ILTN)} that is capable to recognize \emph{if} and
\emph{when} a set of events $\epsilon_1,\dots,\epsilon_k$ happens in
the sequence, under the hypothesis, that certain constraints
expressed in IRL are (softly) satisfied. 

We implemented a first simple prototype of ILTN
in TensorFlow as a wrapper of LTN~\citep{LTN}.
The code is available at \url{https://github.com/sbadredd/interval-ltn}.
The present
section describes important design choices that enable the
architecture. In the description, we concentrate only on the
temporal prediction and not on the classification of the events.

\subsection{Neural Architecture}
\begin{figure}
    \centering
    \begin{subfigure}[b]{0.2\linewidth}
        \centering
        \includegraphics[width=1.\textwidth]{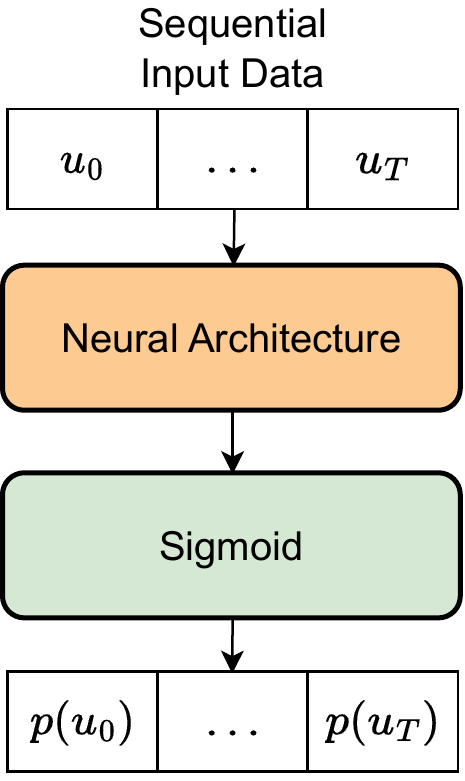}
        \caption{$p(\bm u)$}
        \label{fig:grounding_pred}
    \end{subfigure}
    \qquad\qquad
    \begin{subfigure}[b]{0.36\linewidth}
        \centering
        \includegraphics[width=1.\textwidth]{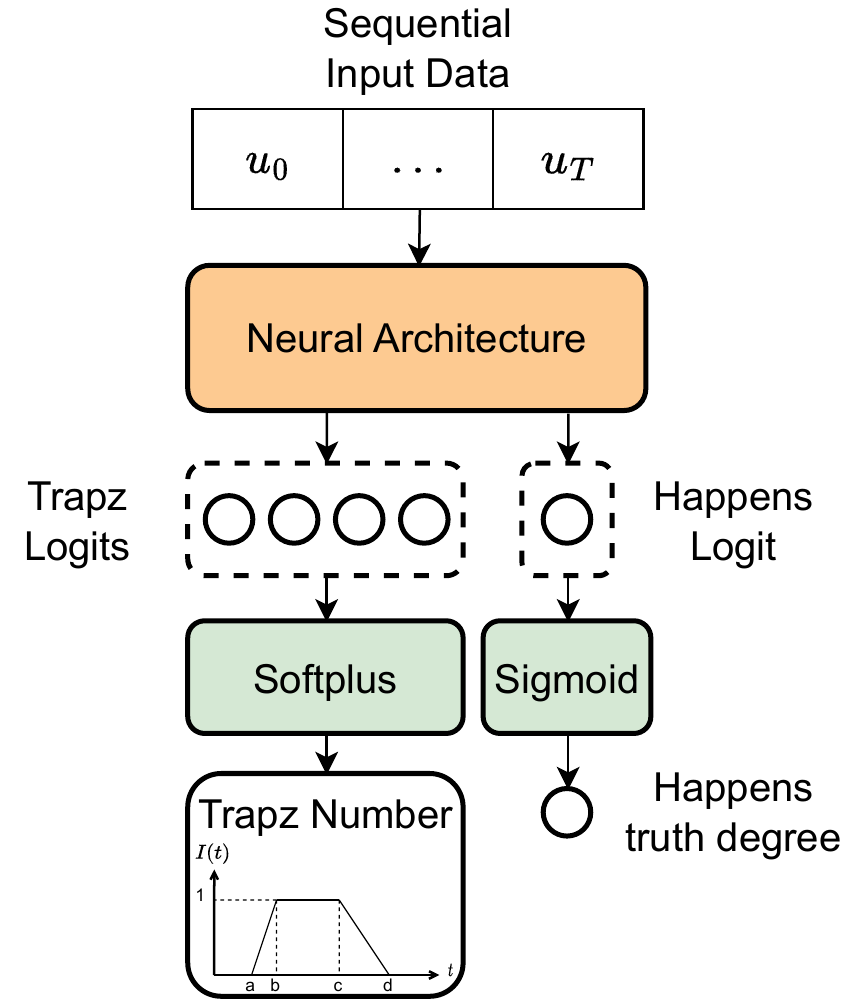}
        \caption{$\epsilon(\bm u)$}
        \label{fig:grounding_event}
    \end{subfigure}
    \caption{Implementation of a temporal predicate symbol from $\L_t$ (left) and of an event symbol (fuzzy interval and happening predicate) from $\L_e$ (right).
    Examples of sequential neural architectures are Recurrent Neural Networks or Transformers.}
    \label{fig:grounding}
\end{figure}

Figure \ref{fig:grounding_event} illustrates how neural networks are
used to ground any event $\epsilon$.  An event is
characterized by two elements: a truth degree
$\Happ(\epsilon)$ indicating if the event happens, and by a
trapezoid interval defining the membership function and when it
happens.

Let us call \emph{logits} the vector of raw (non-normalized) predictions output by the neural model, as is common in the machine learning literature.
The truth degree $\Happ(\epsilon)$
is easily implemented using a single logit node which is then passed to a sigmoid normalization function and constrained in the interval $[0,1]$.

Directly defining the parameters $(a,b,c,d)$ 
of the interval is difficult as the semantic constraint $a \leq b \leq c \leq d$ is hard to implement in a neural architecture.
Instead, our neural architecture predicts the four values $(a,b-a,c-b,d-c)$.
The only semantical constraint on these values is that each is positive.
This is easily implemented using four logit nodes which are then passed through softplus activations.

\subsection{Smooth Membership Functions}
\newcommand{\smooth}{\sim}
We notice an important vanishing gradient issue with trapezoidal interval functions.
If $x$ is in the flat regions $I(x)=0$ or $I(x)=1$, then $\pdv{I(x)}{x} = 0$.
To account for this,
we define $I_\smooth$, a smooth version of the membership function \eqref{eq:fuzzy-interval}:
\begin{align}
\label{eq:smooth-fuzzy-interval}
I_\smooth(x) & = \begin{cases}
            \softplus(x-a)  & \mbox{if $x \leq a$,} \\
            \softplus(\max(b-x, x-c)) & \mbox{if $b < x \leq c$,} \\
            \softplus(d-x) & \mbox{if $d<x$,} \\
            I(x) & \mbox{otherwise.}
           \end{cases}
\end{align}

Where $\softplus$ is the softplus function defined by:
\begin{align}
    \softplus(x \mid \beta) &= \frac{1}{\beta}\log\left( 1 + e^{\beta x} \right)\\
    \label{eq:pdv_softplus}
    \pdv{\softplus(x \mid \beta)}{x} &= \frac{1}{1+\exp(-\beta x)}
\end{align}

\begin{figure}
    \centering
    \begin{subfigure}[t]{0.4\linewidth}
        \centering
        \begin{tikzpicture}
            \begin{axis}[domain=-3:3,legend pos = north west,width=0.85\textwidth, 
                legend style={nodes={scale=0.7, transform shape}}]
            \addplot[color=blue,mark=None,samples=100,domain=-4:4]{(1/1.) * ln( 1+exp(1.*x))};
            \addlegendentry{$\beta=1$}            
            \addplot[color=red,mark=None,samples=100,domain=-4:4]{(1/2.) * ln( 1+exp(2.*x))};
            \addlegendentry{$\beta=2$}
            \addplot[color=orange,mark=None,samples=100,domain=-4:4]{(1/0.5) * ln( 1+exp(0.5*x))};
            \addlegendentry{$\beta=0.5$}
            \end{axis}
            \end{tikzpicture}
        \caption{$\softplus(x\mid \beta)$}
    \end{subfigure}%
    ~ 
    \begin{subfigure}[t]{0.4\linewidth}
        \centering
        \begin{tikzpicture}
        \begin{axis}[legend pos = north west, width=0.85\textwidth]
            \addplot[color=blue,mark=None,samples=100,domain=-4:4]{1/(1+exp(-(1.*x)))};
            \addplot[color=red,mark=None,samples=100,domain=-4:4]{1/(1+exp(-(2.*x)))};
            \addplot[color=orange,mark=None,samples=100,domain=-4:4]{1/(1+exp(-(0.75*x)))};
        \end{axis}
        \end{tikzpicture}
        \caption{$\pdv{\softplus(x\mid \beta)}{x}$}
    \end{subfigure}
    \caption{The softplus function.}
    \label{f:softplus}
    \end{figure}
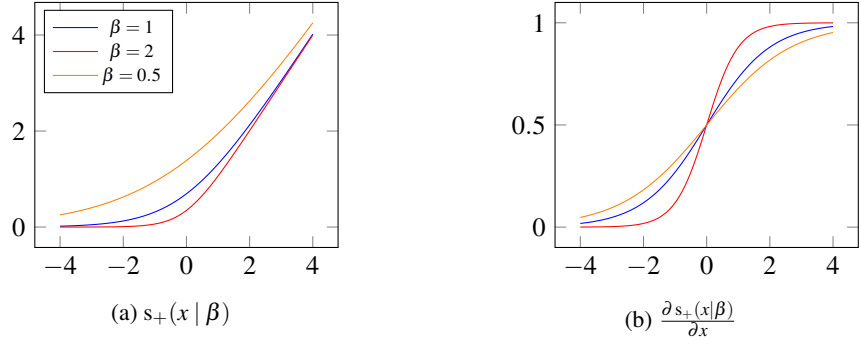

Notice that, in \eqref{eq:smooth-fuzzy-interval}, the inputs to the $\softplus$ function are all negative values.
Intuitively, looking at the graph of softplus in Figure \ref{f:softplus}, $\softplus$ applied to negative values outputs a value that tends to zero with non-negative gradients.

\begin{figure}
    \centering
    \includegraphics[width=0.6\linewidth]{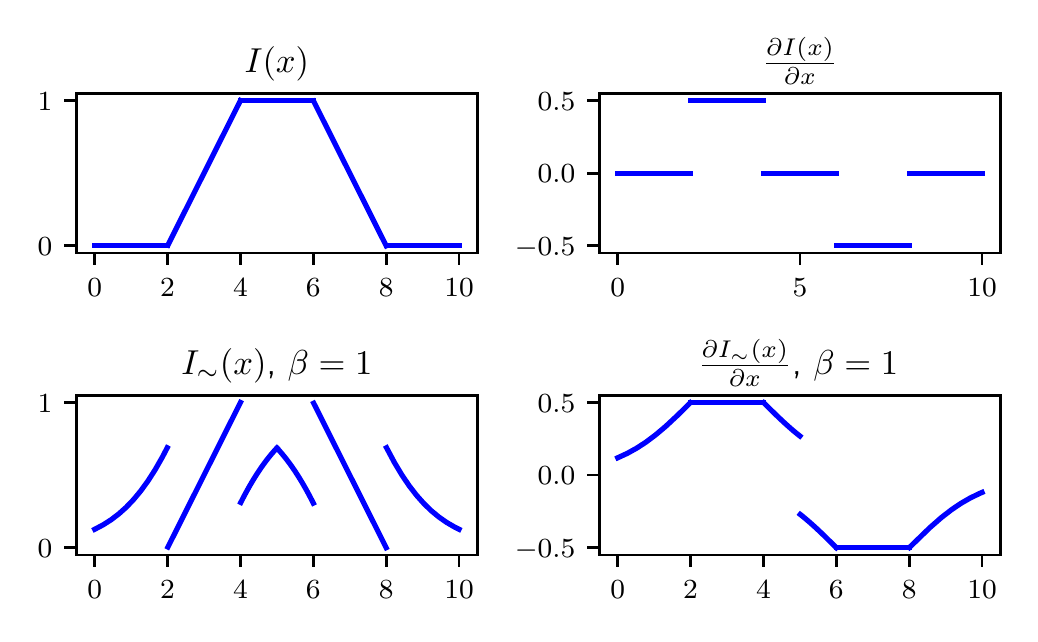}
    \caption{Smooth membership function. The forward pass uses $I(x)$ (top left).
    The backward pass uses $\pdv{I_\smooth(x)}{x}$ (bottom right).}
    \label{fig:mf_func}
\end{figure}

We use $I$ and $I_\smooth$ to define an artificial operator with distinct properties in the forward pass and backward pass of the computational graph.
\footnote{See also \url{https://www.tensorflow.org/api_docs/python/tf/custom_gradient}.}
Let $\epsilon=e(t_1,\dots,t_m)$ be an event term associated with an interval $I$ and a corresponding smooth version $I_\smooth$.
We use:
\begin{align}
    \epsilon(x) & = I(x) \\
    \pdv{\epsilon(x)}{x} &= \pdv{I_\smooth(x)}{x}
\end{align}
The motivation is demonstrated in Figure \ref{fig:mf_func}. 
The backward pass $\pdv{I_\smooth(x)}{x}$ has non-zero gradients everywhere that push $x$ to fit in the center of the interval.
The forward pass remains the accurate evaluation $I(x)$.

Finally, we use the parameter $\beta$ to ensure the accuracy of the operator.
For example, for large negative differences $x-a$, the output of $\softplus(x-a)$ gets very small 
and can become zero because of the way computers approximate real numbers.
In float32 precision format, this happens with $x-a>90$ and $\beta=1$.
In such cases, the gradients still vanish.
We avoid this issue by setting $\beta=\frac{1}{T}$, where $T$ is the largest time difference occuring in our data, 
or in other words $T$ is the length of the trace in the experiment.

\subsection{Smooth Relations}
Let $A=(a,b,c,d)$ and $B=(a',b',c',d')$ be two trapezoids. 
Without loss of generality, suppose that $a \leq a'$.
Similarly to how membership functions has zero gradients on some parts of the domain, the relations in \ref{s:relations} 
have vanishing gradients in two situations.
The first is when $A \contained B = 0$ and the trapezoids do not intersect. 
In other words, when $d<a'$.
The second is when $A \contained B = 1$ and $A$ is fully contained in $B$.
In other words, when $a>a'$, $b>b'$, $c<c'$, and $d<d'$.

Again, we solve this by defining a smooth operator for the backward pass relying on the softplus operator:
\begin{equation}
    (A \contained B)_{\smooth} = \begin{cases}
        \softplus(d-a)  & \mbox{if $d < a'$,} \\
        \softplus(a'-a+d-d')  & \mbox{if $A$ fully in $B$,} \\
        A \contained B & \mbox{otherwise.}
    \end{cases}
\end{equation}
with the non-vanishing derivatives on the trapezoid edges of $A$ and $B$. 
We use $A \contained B$ in the forward pass of the computational graph and $\pdv{(A \contained B)_\smooth}{x}$ in the backward pass, where $x$ is any parameter defining $A$ or $B$.
Finally, we still set $\beta=\frac{1}{T}$ for the softplus function.

\begin{table*}[h]
    \centering
    \begin{tabular}{ m{1cm}  c  m{3cm}  m{3cm}  c  }
      \toprule
      Task & Initial Conditions & Setting & Constraints & Result \\ 
      \midrule
      T1 &
      \begin{minipage}{.2\textwidth}
        \includegraphics[width=\linewidth]{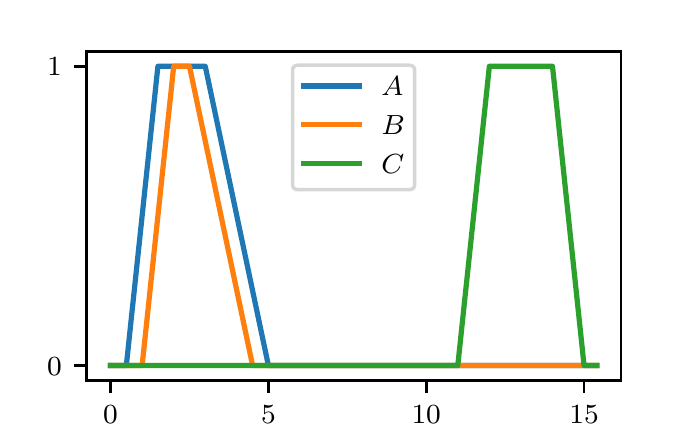}
      \end{minipage}
      &
        \begin{itemize}[topsep=0pt,leftmargin=5.5mm]
          \item $B$ trainable,
          \item $A$ and $C$ fixed.
        \end{itemize}
      & 
        \begin{enumerate}[topsep=0pt,leftmargin=5.5mm]
            \item $|B| \approx 2$
            \item $B \after A$
            \item $B \before C$
        \end{enumerate}
      &
      \begin{minipage}{.2\textwidth}
        \includegraphics[width=\linewidth]{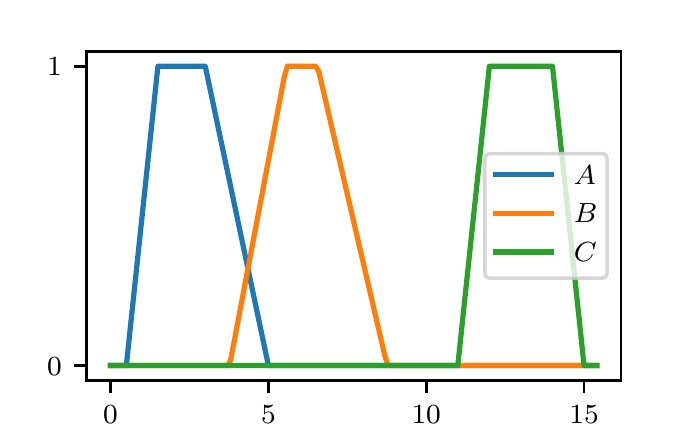}
      \end{minipage} \\ 
      T2 &
      \begin{minipage}{.2\textwidth}
        \includegraphics[width=\linewidth]{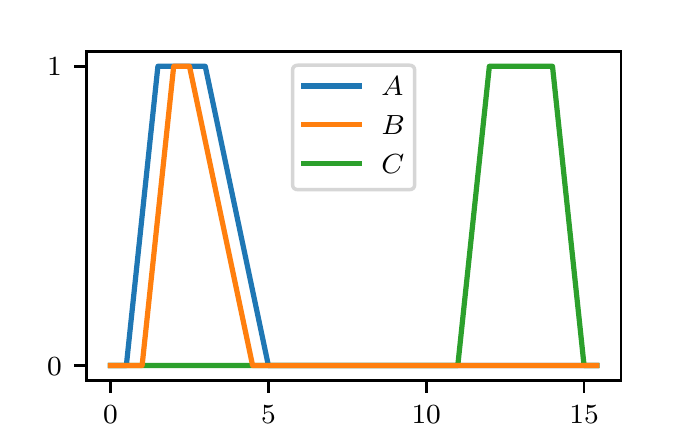}
      \end{minipage}
      &
        \begin{itemize}[topsep=0pt,leftmargin=5.5mm]
          \item $B$ trainable,
          \item $A$ and $C$ fixed.
        \end{itemize}
      & 
        \begin{enumerate}[topsep=0pt,leftmargin=5.5mm]
            \item $|B| \approx 1.5$
            \item $B \starts C$
        \end{enumerate}
      &
      \begin{minipage}{.2\textwidth}
        \includegraphics[width=\linewidth]{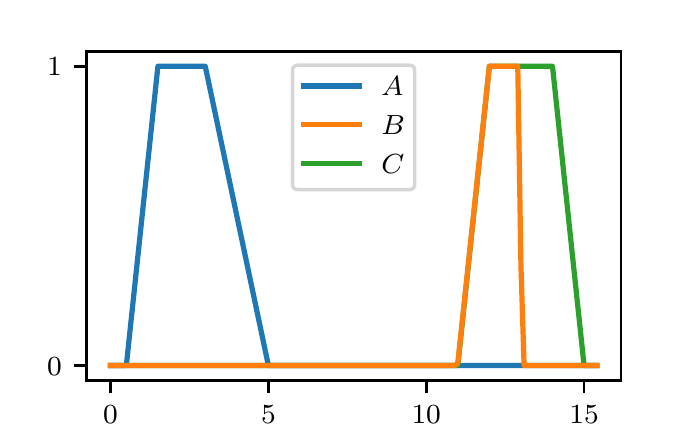}
      \end{minipage} \\ 
      T3 &
      \begin{minipage}{.2\textwidth}
        \includegraphics[width=\linewidth]{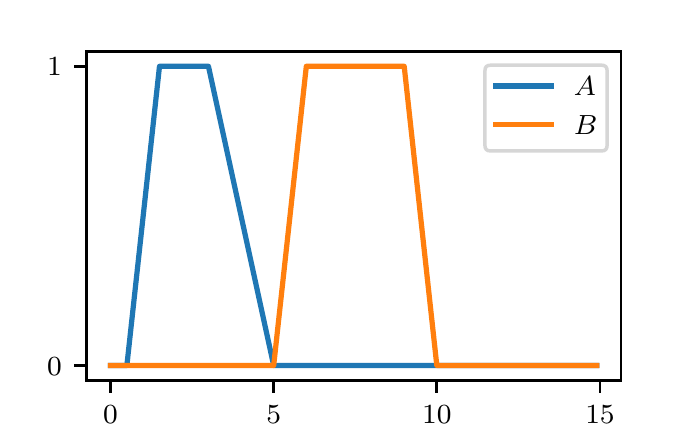}
      \end{minipage}
      &
        \begin{itemize}[topsep=0pt,leftmargin=5.5mm]
          \item $A$ trainable,
          \item $B$ fixed.
        \end{itemize}
      & 
        \begin{enumerate}[topsep=0pt,leftmargin=5.5mm]
            \item $A \overlaps B$
            \item $A(3)$
            \item $\lnot A(2)$
        \end{enumerate}
      &
      \begin{minipage}{.2\textwidth}
        \includegraphics[width=\linewidth]{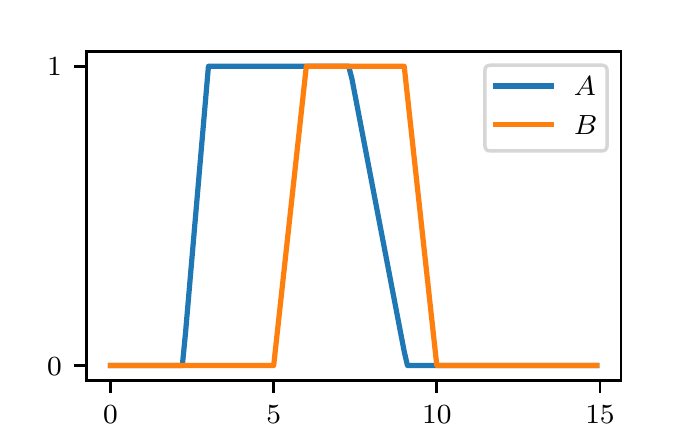}
      \end{minipage} \\ 
      T4 &
      \begin{minipage}{.2\textwidth}
        \includegraphics[width=\linewidth]{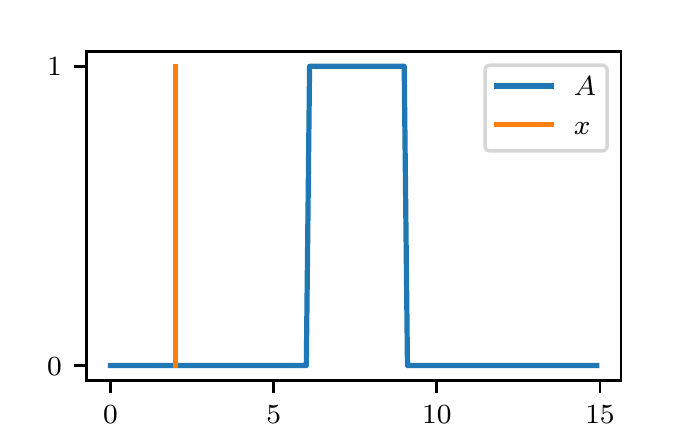}
      \end{minipage}
      &
        \begin{itemize}[topsep=0pt,leftmargin=5.5mm]
          \item $x$ trainable,
          \item $A$ fixed.
        \end{itemize}
      & 
        \begin{enumerate}[topsep=0pt,leftmargin=5.5mm]
            \item $\End(A)(x)$
        \end{enumerate}
      &
      \begin{minipage}{.2\textwidth}
        \includegraphics[width=\linewidth]{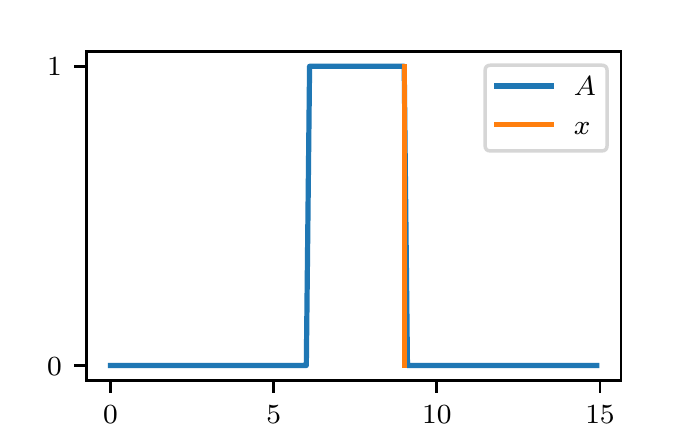}
      \end{minipage} \\

      \bottomrule
    \end{tabular}
    \caption{Experiments}
    \label{tab:experiments}
    
\end{table*}

\section{Experiments}
\label{sec:experiments}

We test the system on synthetic tasks that require a combination of learning and reasoning about temporal relations between fuzzy intervals.
Let $\phi_1, \dots, \phi_n$ be $n$ constraints written in Interval Real Logic defining a knowledge base $\mathcal{K}$. 
Like in LTN~\citep{LTN}, the grounding of the knowledge base defines a satisfaction level to maximise.
Optimising by gradient descent, we have the following loss function:
\begin{equation}
    L(\mathcal{K}, \theta) - (\G(\phi_1, \theta) \land \dots \land \G(\phi_n , \theta))
\end{equation}
where $\phi_i$'s are $\L_e$ formulas and $\theta$ is a set of trainable parameters used to define the grounding. 
We focus the experimental study on the training of events with constraints written using $\L_e$, which is the main innovation of this paper.
Specifically, we focus on learning parameters that define the fuzzy trapezoid intervals of events. 

Table \ref{tab:experiments} displays a list of training experiments where ILTN maximizes the satisfaction of temporal constraints.
All tasks are trained using the Adam optimizer~\citep{Adam} with a learning rate of $0.1$.
In T1, T2, T3, and T4, the results are obtained after training for 50, 500, 5000, and 200 training steps, respectively. 
For the logical operators, we use the product t-norm $u \land v = u v$ and the standard negation $\lnot u = 1 - u$. 

We highlight the following features:
\begin{itemize}
    \item In T1, T2 and T3, the system learns fuzzy intervals,
    \item In T4, the system learns a time point value $x$,
    \item T1, T2, and T3 display constraints using Allen's relational symbols $\after$, $\before$, $\starts$, and $\overlaps$,
    \item T3 and T4 display constraints using membership functions,
    \item T4 displays a constraint using a functional symbol $\End$,
    \item In T1 and T2, $u \approx v$ is a smooth equality predicate implemented as $\exp(-|u-v|) \in [0,1]$.
\end{itemize}

\subsection{Challenges for Future Work}
\label{s:future_work}
In all tasks, the system learns to update the event groundings to satisfy the knowledge base.
The experiments demonstrate that ILTN can successively backpropagate gradients through the Interval Real logic.
Nevertheless, we highlight three limitations of our experiments that future work should explore.

Firstly, early stopping was an important factor in our experiments.
Continuing training after reaching the maximal satisfaction level could sometimes lead to worsening the results.
This is likely due to the smoothing of operators for obtaining non-vanishing gradients.
This feature is important early in training.
However, once a constraint is satisfied, having vanishing gradients is acceptable. 
Future work could explore reducing or stopping the smoothening of constraints when their satisfaction levels are high. 

Secondly, the Adam optimizer is traditionally used with learning rates in the order of $0.001$.
In comparison, the learning rate of $0.1$ used to train our synthetic tasks is unusually high.
A lower learning rate led to experiments not converging fast enough.
There is likely a scaling issue in the gradients of some operations.
This could also explain why T3 required more training steps than the other tasks to reach convergence:
this is the only task that mixes relational operators ($\overlaps$) and membership functions. 
The two have gradients scaling differently which can challenge the training.
Future work should analyse further the gradient properties of each temporal operator.

Thirdly, the present experiments do not showcase yet the power of learning events that depend on input features.
For example, in Figure \ref{fig:grounding_event}, the present tasks only learn trapezoid logits that define a trapezoid number.
There is no sequential data in input and neural architecture that builds on top of it.
Future work should explore more elaborate tasks employing such architectures.

\section{Conclusions}
\label{sec:conclusion}

In this paper, we introduce Interval Real Logic (IRL), a two-sorted logic that enables the prediction of properties that evolve within a set of data sequences (traces) and properties of events that occur within the sequences. 
IRL semantics are defined in terms of sequences of real feature vectors, and connectives and quantifiers are interpreted using fuzzy logic. 
We represent event duration through trapezoidal fuzzy intervals, and fuzzy temporal relations are defined based on the relationships between the intervals' areas and their intersections. 

We also present Interval Logic Tensor Networks (ILTN), a neuro-symbolic system that leverages background knowledge expressed in IRL to predict the fuzzy duration of events. 
To prevent vanishing gradient during learning, we use softplus functions to smooth both events and their relations. 
We evaluate ILTN's performance on four tasks with different temporal constraints and show that it is capable of making events compliant with background knowledge in all four tasks.

In Section \ref{s:future_work}, we suggest several directions for future research. 
One promising avenue would be to test ILTN on more realistic and complex scenarios, such as those involving real-world data. 
Scalability is a significant challenge for neuro-symbolic frameworks for event recognition, which we have hoped to address by representing events as a whole, rather than using a point-wise approach.

\bibliographystyle{unsrtnat}
\bibliography{biblio}
\end{document}